\documentclass[letterpaper, 10 pt, conference]{ieeeconf} 

\IEEEoverridecommandlockouts                              

\usepackage{multicol}
\usepackage{outlines}

\usepackage{url}
\usepackage{graphicx}
\usepackage{amsmath}
\usepackage{amsfonts}
\usepackage{makecell}
\usepackage{float}
\usepackage{outlines}
\usepackage{sidecap}
\usepackage{siunitx}
\usepackage[colorlinks=true]{hyperref}
\hypersetup{linkcolor=black}
\hypersetup{citecolor=black}
\usepackage{flushend}
\usepackage[sorting=none, natbib=true, style=numeric-comp]{biblatex}
\newcommand{\norm}[1]{\left\lVert#1\right\rVert}

\newcommand{\lhat}[0]{\mathbf{l}}
\newcommand{\m}[0]{\mathbf{m}}
\newcommand{\bm}[1]{\mathbf{#1}}

\begin{document}

\title{ScrewNet: Category-Independent Articulation Model Estimation From Depth Images Using Screw Theory}


 \author{Ajinkya Jain$^{1}$, Rudolf Lioutikov$^{1}$, Caleb Chuck$^{1}$ and Scott Niekum$^{1}$%
\thanks{$^{1}$Personal Autonomous Robotics Lab (PeARL), The University of Texas at Austin. Contact: {\tt\small ajinkya@utexas.edu}}}%

\maketitle


\begin{abstract}
Robots in human environments will need to interact with a wide variety of articulated objects such as cabinets, drawers, and dishwashers while assisting humans in performing day-to-day tasks. Existing methods either require objects to be textured or need to know the articulation model category a priori for estimating the model parameters for an articulated object. We propose ScrewNet, a novel approach that estimates an object's articulation model directly from depth images without requiring a priori knowledge of the articulation model category. ScrewNet uses screw theory to unify the representation of different articulation types and perform category-independent articulation model estimation. We evaluate our approach on two benchmarking datasets and three real-world objects and compare its performance with a current state-of-the-art method. Results demonstrate that ScrewNet can successfully estimate the articulation models and their parameters for novel objects across articulation model categories with better on average accuracy than the prior state-of-the-art method. Project page: \href{https://pearl-utexas.github.io/ScrewNet/}{https://pearl-utexas.github.io/ScrewNet/}
\end{abstract}


\section{Introduction}
\label{sec:intro}
Human environments are populated with objects that contain functional parts, such as refrigerators, drawers, and staplers. These objects are known as articulated objects and consist of multiple rigid bodies connected via mechanical joints such as hinge joints or slider joints. Service robots will need to interact with these objects frequently. For manipulating such objects safely, a robot must reason about the articulation properties of the object. Safe manipulation policies for these interactions can be obtained directly either by using expert-defined control policies \cite{jain2009pulling, baum2017opening} or by learning them through interactions with the objects \cite{gupta2019relay, kroemer2019review}. However, this approach may fail to provide good manipulation policies for all articulated objects that the robot might interact with, due to the vast diversity of articulated objects in human environments and the limited availability of interaction time. An alternative is to estimate the articulation models through observations, and then use a planning \cite{jain2018efficient} or model-based RL method \cite{kroemer2019review} to manipulate them effectively.

Existing methods for estimating articulation models of objects from visual data either use fiducial markers to track the relative movement between the object parts \cite{sturm2011probabilistic, niekum2015online, liu2019learning} or require textured objects so that feature tracking techniques can be used to observe this motion \cite{pillai2015learning, martin2019coupled, jain2019learning}. These requirements severely restrict the class of objects on which these methods can be used. Alternatively deep networks can extract relevant features from raw images automatically for model estimation \cite{abbatematteo2019learning, li2020category}. However, these methods assume prior knowledge of the articulation model category (revolute or prismatic) to estimate the category-specific model parameters, which may not be readily available for novel objects encountered by robots in human environments. Addressing this limitation, we propose a novel approach, ScrewNet, which uses screw theory to perform articulation model estimation directly from depth images without requiring prior knowledge of the articulation model category. ScrewNet unifies the representation of different articulation categories by leveraging the fact that the common articulation model categories (namely revolute, prismatic, and rigid) can be seen as specific instantiations of a general constrained relative motion between two objects about a fixed screw axis. This unified representation enables ScrewNet to estimate the object articulation models independent of the model category.

\begin{figure}
    \centering
    \includegraphics[width=0.42\textwidth]{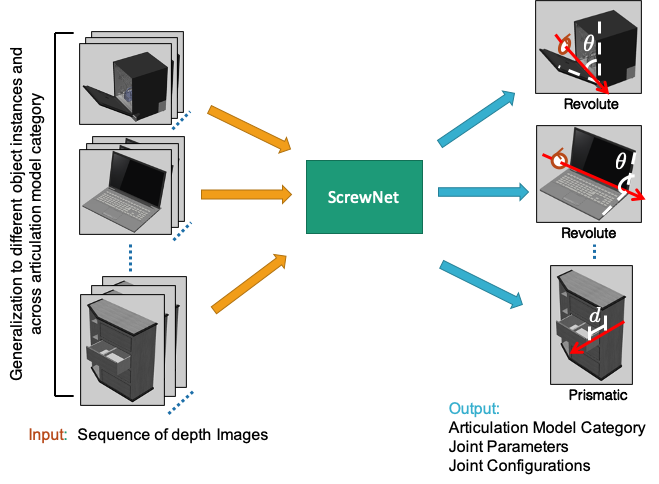}
    \vspace*{-3.0ex}
  \caption{ScrewNet estimates the articulation model for objects directly from depth images and can generalize to novel objects within and across articulation model categories}
  \vspace*{-3ex}
\end{figure}

ScrewNet garners numerous benefits over existing approaches. First, it can estimate articulation models directly from raw depth images without requiring a priori knowledge of the articulation model category. Second, due to the screw theory priors, a single network suffices for estimating models for all common articulation model categories unlike prior methods \cite{abbatematteo2019learning, li2020category}. Third, ScrewNet can also estimate an additional articulation model category, the helical model (motion of a screw), without making any changes in the network architecture or the training procedure.

We conduct a series of experiments on two benchmarking datasets: a simulated articulated objects dataset \cite{abbatematteo2019learning}, and the PartNet-Mobility dataset \cite{Xiang_2020_SAPIEN, Mo_2019_CVPR, chang2015shapenet}, and three real-world objects: a microwave, a drawer, and a toaster oven. We compare ScrewNet with a current state-of-the-art method \cite{abbatematteo2019learning} and three ablated versions of ScrewNet and show that it outperforms all baselines with a significant margin.

\section{Related Work}
\label{sec:litsurvey}
\textbf{Articulation model estimation from visual observations:} \citet{sturm2009learning} proposed a probabilistic framework to learn the articulation relationships between different parts of an articulated object from the time-series observations of 6D poses of object parts \cite{sturm2009learning, sturm2011probabilistic}. The framework was extended to estimate the articulation model for textured objects directly from raw RGB images by extracting SURF features from the images and tracking them robustly\cite{pillai2015learning}. Other approaches have explored modeling articulated objects that exhibit configuration-dependent changes in the articulation model, rather than having a single model throughout their motion \cite{niekum2015online, jain2019learning}. Recently, the problem of articulation model parameter estimation was posed as a regression task given a known articulation model category \cite{abbatematteo2019learning}. The mixture density network-based approach predicts model parameters using a single depth image.
However, in a realistic setting, an object's articulation model category might not be available a priori to the robot.

\textbf{Interactive perception (IP):} IP approaches leverage the robot's interaction with the objects for generating a rich perceptual signal for robust articulation model estimation \cite{ martin2014online, hausman2015active, bohg2017interactive}. \citet{katz2008manipulating} first studied IP to learn articulated motion models for planar objects \cite{katz2008manipulating}, and later extended it to learn 3D kinematics of articulated objects \cite{katz2013interactive}. In more recent works, \cite{martin2016integrated} and \cite{martin2019coupled} have further extended the approach and used hierarchical recursive Bayesian filters to develop online algorithms from articulation model estimation from RGB images. However, current IP approaches still require textured objects for estimating the object articulation model from raw images, whereas, ScrewNet imposes no such requirement on the objects.

\textbf{Articulated object pose estimation:} For known articulated objects, the problem of articulation model parameter estimation can also be treated as an articulated object pose estimation problem. Different approaches leveraging object CAD model information \cite{michel2015pose, desingh2019factored} and the knowledge of articulation model category \cite{abbatematteo2019learning, Yi2018, li2020category} have been proposed to estimate the 6D pose of the articulated object in the scene. These approaches can be combined with an object detection method, such as YOLOv4 \cite{bochkovskiy2020yolov4}, to develop a pipeline for estimating the articulation model parameters for objects from raw images. On the other hand, ScrewNet can directly estimate the articulation model for an object from depth images without requiring any prior knowledge about it.

\textbf{Other approaches}:
\citet{perez2017c} and \citet{liu2019learning} have proposed methods to learn articulation models as geometric constraints encountered in a manipulation task from non-expert human demonstrations. Leveraging natural language descriptions during demonstrations, Daniele et al. \cite{daniele2020multiview} have proposed a multimodal learning framework that incorporates both vision and natural language information for articulation model estimation. However, these approaches \citep{perez2017c, liu2019learning} use fiducial markers to track the movement of the object, unlike ScrewNet, that works on raw images.

\section{Background}
\label{sec:prelim}
\textbf{Screw displacements:} Chasles' theorem states that \textit{``Any displacement of a body in space can be accomplished by means of a rotation of the body about a unique line in space accompanied by a translation of the body parallel to that line"} \cite{siciliano2016springer}. This line is called the screw axis of displacement, $\mathsf{S}$ \cite{mason2001mechanics, jia2019}. We use Pl\"{u}cker coordinates to represent this line. The Pl\"{u}cker coordinates of the line $l = \mathbf{p} + x\lhat$
are defined as $(\lhat, \m)$, with moment vector $\mathbf{m} = \mathbf{p} \times \lhat$ \cite{mason2001mechanics, jia2019}. The constraints $\norm{\lhat} = 1$ and $\langle\lhat,\m\rangle = 0$ ensure that the degrees of freedom of the line in space are restricted to four. The rigid body displacement in $SE(3)$ is defined as $ \sigma = (\mathbf{l}, \mathbf{m}, \theta, d)$. The linear displacement $d$ and the rotation $\theta$ are connected through the pitch $h$ of the screw axis, $d = h \theta$. The distance between $l_1 := (\lhat_1, \m_1)$ and $l_2 := (\lhat_2, \m_2)$ is defined as:
\begin{equation}
\mathsf{d}((\lhat_1, \m_1), (\lhat_2, \m_2))  = 
\begin{cases}
    0, ~~\text{if $l_1$ and $l_2$ intersect} \\
    \norm{\mathbf{l}_{1} \times (\mathbf{m}_{1} - \mathbf{m}_{2})},~ \text{elif $l_1 \parallel l_2$}\\[0.5px]
	\dfrac{\lvert \lhat_1 \cdot \m_2 + \lhat_2 \cdot \m_1 \rvert} {\norm{\mathbf{l}_{1} \times \mathbf{l}_{2}}},~ \text{else} 
\end{cases}
\label{eq:1}
\end{equation}

\textbf{Frame transformations on Pl\"{u}cker lines:}
Given a rotation matrix $R$ and a translation vector $\mathbf{t}$ between two frames $\mathcal{F}_A$ and $\mathcal{F}_B$, a 3D line displacement matrix $\tilde{D}$ can be defined between the two frames for transforming a line $l := (\lhat, \m)$ from frame $\mathcal{F}_A$ to frame $\mathcal{F}_B$ as: 
\begin{equation}
    \begin{gathered}
        \begin{bmatrix}
        ^B\lhat \\ ^B\m
        \end{bmatrix} = ~^B\tilde{D}_A ~ \begin{bmatrix} ^A\lhat \\ ^A\m \end{bmatrix}, \\ \text{where,} ^B\tilde{D}_A = 
        \begin{bmatrix}
            R & \mathbf{0} \\
            [\mathbf{t}]_{\times}R & R
        \end{bmatrix},
        [\mathbf{t}]_{\times} = 
        \begin{bmatrix}
            0 & -t_3 & t_2 \\
            t_3 & 0 & -t_1 \\
            -t_2 & t_1 & 0
        \end{bmatrix}
        \label{eq:2}
    \end{gathered}
\end{equation}
where $[\mathbf{t}]_{\times}$ denotes the skew-symmetric matrix corresponding to the translation vector $\mathbf{t}$, and $(^A\lhat, ^A\m)$ and $(^B\lhat, ^B\m)$ represents the line $l$ in frames $\mathcal{F}_A$ and $\mathcal{F}_B$, respectively \cite{bartoli20013d}.

\section{Approach}
\label{sec:method}
\subsection{Problem Formulation}
\label{sec:problem}
Given a sequence of $n$ depth images $\mathcal{I}_{1:n}$ of motion between two parts of an articulated object, we wish to estimate the articulation model $\mathcal{M}$ and its parameters $\phi$ governing the motion between the two parts without knowing the articulation model category a priori. Additionally, we wish to estimate the configurations $q_{1:n}$ that uniquely identify different relative spatial displacements between the two parts in the given sequence of images $\mathcal{I}_{1:n}$ under model $\mathcal{M}$ with parameters $\phi$. We consider articulation models with at most one degree-of-freedom (DoF), i.e. $\mathcal{M} \in \{\mathcal{M}_{\text{rigid}}, \mathcal{M}_{\text{revolute}}, \mathcal{M}_{\text{prismatic}}, \mathcal{M}_{\text{helical}}\}$. Model parameters $\phi$ are defined as the parameters of the screw axis of motion, i.e. $\mathsf{S} = (\lhat, \m)$, where both $\lhat$ and $\m$ are three-dimensional real vectors. Each configuration $q_i$ corresponds to a tuple of two scalars, $q_i = (\theta_i, d_i)$, defining a rotation around and a displacement along the screw axis $\mathsf{S}$. We assume that the relative motion between the two object parts is governed only by a single articulation model. 

\subsection{ScrewNet}
We propose ScrewNet, a novel approach that given a sequence of segmented depth images $\mathcal{I}_{1:n}$ of the relative motion between two rigid objects estimates the articulation model $\mathcal{M}$ between the objects, its parameters $\phi$, and the corresponding configurations $q_{1:n}$. In contrast to  state-of-the-art approaches, ScrewNet does not require a priori knowledge of the articulation model category for the objects to estimate their models. ScrewNet achieves category independent articulation model estimation by representing different articulation models through a unified representation based on the screw theory~\cite{siciliano2016springer}.
ScrewNet represents the 1-DoF articulation relationships between rigid objects (rigid, revolute, prismatic, and helical) as a sequence of screw displacements along a common screw axis. A rigid model is now defined as a sequence of identity transformations, i.e., $\theta_{1:n}=0 \wedge d_{1:n}=0$, a revolute model as a sequence of pure rotations around a common axis, i.e., $\theta_{1:n} \neq 0 \wedge d_{1:n}=0$, a prismatic model as a sequence of pure displacements along the same axis, i.e., $\theta_{1:n} = 0 \wedge d_{1:n} \neq 0$, and, a helical model as a sequence of correlated rotations and displacements along a shared axis, i.e., $\theta_{1:n} \neq 0 \wedge d_{1:n} \neq 0$).

Under this unified representation, all 1-DoF articulation models can be represented using the same number of parameters, i.e., 6 parameters for the common screw axis $\mathsf{S}$ and $2n = |\{(\theta_i, d_i) \forall i \in \{1...n\} \}|$ parameters for configurations, which enables ScrewNet to perform category independent articulation model estimation. ScrewNet not only estimates the articulation model parameters without requiring the model category $\mathcal{M}$, but is also capable of estimating the category itself. This ability can potentially reduce the number of control parameters required for manipulating the object \cite{jain2019learning}. A unified representation also allows ScrewNet to use a single network to estimate the articulation motion models across categories, unlike prior approaches that required separate networks, one for each articulation model category \cite{abbatematteo2019learning, li2020category}. Having a single network grants ScrewNet two major benefits: first, it needs to train fewer total parameters, and second, it allows for a greater sharing of training data across articulation categories, resulting in a significant increase in the number of training examples that the network can use. Additionally, in theory, ScrewNet can also estimate an additional articulation model category, the helical model, which was not addressed in earlier work \cite{abbatematteo2019learning,  martin2019coupled, sturm2011probabilistic}.

\begin{figure}[t]
    \centering
    \includegraphics[width=0.42\textwidth]{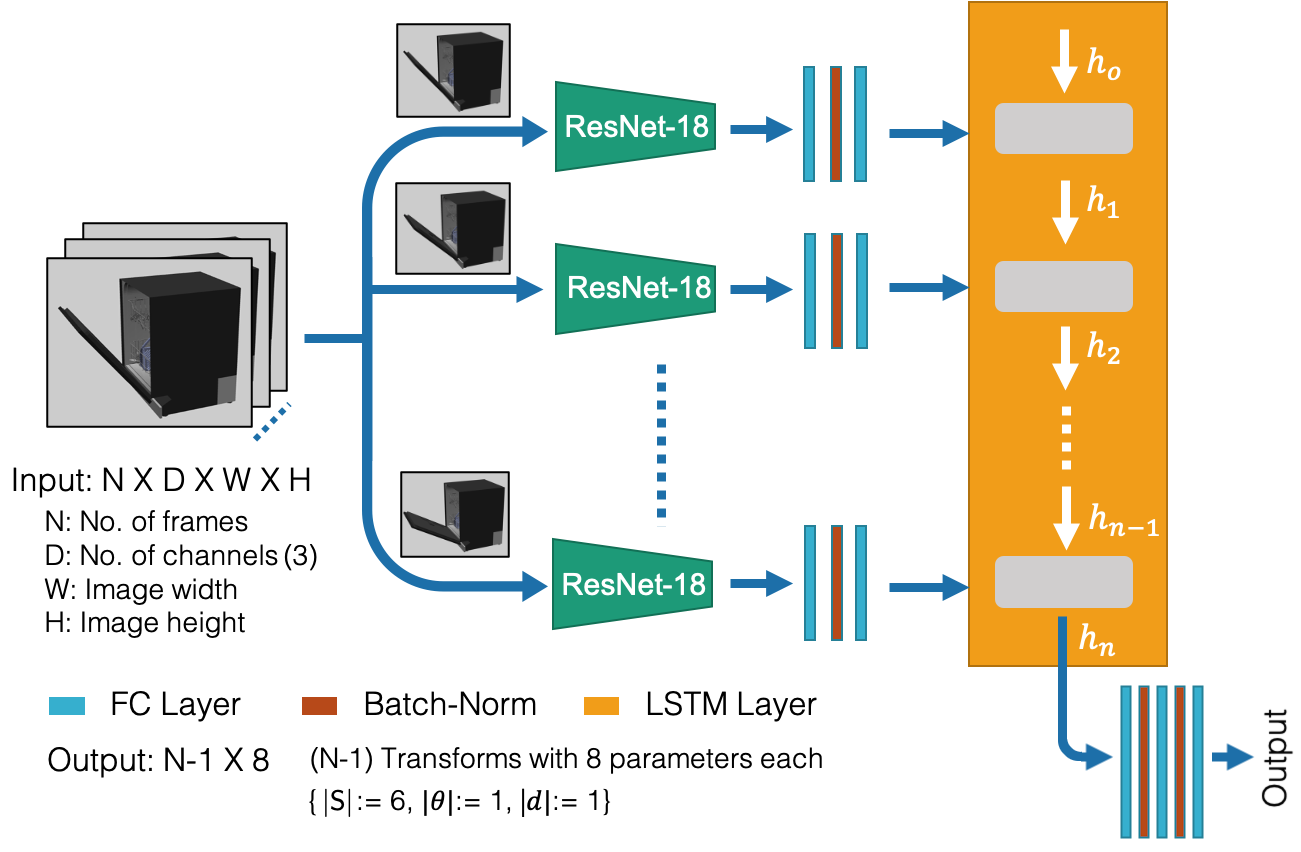}
    \vspace*{-2.0ex}
  \caption{Taking a sequence of depth images as input, ScrewNet first extracts features from the depth images using ResNet, passes them through an LSTM layer to encode their sequential information, and then uses MLP to predict a sequence of screw displacements having a shared screw axis}
  \label{fig:arch}
  \vspace*{-3ex}
\end{figure}

\textbf{Architecture:}
ScrewNet sequentially connects a ResNet-18 CNN \cite{he2016deep}, an LSTM with one hidden layer, and a 3-Layer MLP. ResNet-18 extracts features from the depth images that are fed into the LSTM which encodes the sequential information from the extracted features into a latent representation. Using this representation, the MLP then predicts a sequence of screw displacements having a common screw axis. The network is trained end-to-end, with ReLU activations for the fully-connected layers. Fig.~\ref{fig:arch} shows the network architecture. The model category $\mathcal{M}$ is deduced from the predicted screw displacements using a decision-tree based on the displacements properties of each model class.

\textbf{Loss function:} 
Screw displacements are composed of two major components: the screw axis $\mathsf{S}$, and the corresponding configurations $q_i$ about it. Hence, we pose ScrewNet training as a multi-objective optimization problem with loss
\begin{equation}
    \mathcal{L} = \lambda_{1} \mathcal{L}_{\mathsf{S}_{\text{ori}}} + \lambda_{2} \mathcal{L}_{\mathsf{S}_{\text{dist}}} + \lambda_{3} \mathcal{L}_{\mathsf{S}_{\text{cons}}} + \lambda_4 \mathcal{L}_{\text{q}},
\end{equation}
where $\lambda_i$ weights the respective component. $\mathcal{L}_{\mathsf{S}_{\text{ori}}}$ penalizes the screw axis orientation mismatch as the angular difference between the target and the predicted orientations. $\mathcal{L}_{\mathsf{S}_{\text{dist}}}$ penalizes the spatial distance between the target and predicted screw axes as defined in Eqn.~\ref{eq:1}. $\mathcal{L}_{\mathsf{S}_{\text{cons}}}$ enforces the constraints $\langle\lhat,\m\rangle =0$ and $\norm{\lhat} = 1$. $\mathcal{L}_{\text{q}} := \alpha_1 \mathcal{L}_{\theta} + \alpha_2 \mathcal{L}_{d}$ penalizes errors in the configurations, where $\mathcal{L}_{\theta}$ and $\mathcal{L}_{d}$ represent the rotational and translational error respectively:
\begin{equation}
    \begin{gathered}
        \mathcal{L}_{\theta} =  \mathbf{I}_{3,3}- R(\theta_{\text{tar}}; \lhat_{\text{tar}})~R(\theta_{\text{pred}}; \lhat_{\text{pred}})^T, \\
        \mathcal{L}_d = \norm{d_{\text{tar}}\cdot \lhat_{\text{tar}} - d_{\text{pred}} \cdot \lhat_{\text{pred}}}
    \end{gathered}
\end{equation}
with $R(\theta; \lhat)$ denoting the rotation matrix corresponding to a rotation of angle $\theta$ about the axis $\lhat$. We choose this particular form of the loss function for $\mathcal{L}_{\text{q}}$, rather than a standard loss function such as an $L2$ loss, as it ensures that the network predictions are grounded in their physical meaning. By imposing a loss based on the orthonormal property of the $3D$ rotations, the proposed loss function ensures that the learned angle-axis pair $(\lhat_{\text{pred}}, \theta_{\text{pred}})$ corresponds to a rotation $R(\theta_{\text{tar}}; \lhat_{\text{tar}}) \in SO(3)$. Similarly, the loss function $\mathcal{L}_d$ calculates the difference between the two displacements along two different axes $\lhat_{\text{tar}}$ and $\lhat_{\text{pred}}$, rather than calculating the difference between the two configurations, $d_{\text{tar}}$ and $d_{\text{pred}}$, which assumes that they represent displacements along the same axis. Hence, this choice of loss function ensures that the network predictions conform to the definition of a screw displacement. We empirically choose weights to be $\lambda_1=1, \lambda_2=2, \lambda_3=1, \lambda_4=1, \alpha_1=1$, and $\alpha_2=1$.

\textbf{Training data generation:}
The training data consists of sequences of depth images of objects moving relative to each other and the corresponding screw displacements. The objects and depth images are rendered in Mujoco \cite{todorov2012mujoco}. We apply random frame skipping and pixel dropping to simulate noise encountered in real world sensor data. We use the cabinet, drawer, microwave, the toaster-oven object classes from the simulated articulated object dataset\cite{abbatematteo2019learning}. The cabinet, microwave, and toaster object classes contain a revolute joint each, while the drawer class contains a prismatic joint. We consider both left-opening cabinets and right-opening cabinets. From the PartNet-Mobility dataset \cite{Xiang_2020_SAPIEN, Mo_2019_CVPR, chang2015shapenet}, we consider the dishwasher, oven, and microwave object classes for the revolute articulation model category, and the storage furniture object class consisting of either a single column of drawers or multiple columns of drawers, for the prismatic articulation model category.

\begin{SCfigure}
    \centering
  \includegraphics[width=0.25\textwidth]{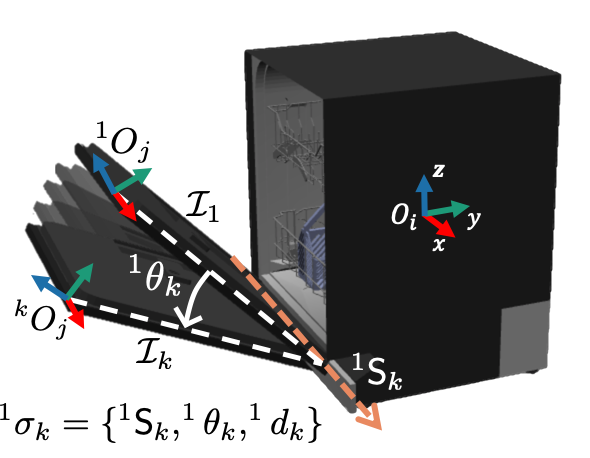}
  \caption{The training labels are generated by calculating the screw displacements between the temporally displaced poses of the object $o_j$, and expressing them in a frame of reference attached to the base object $o_i$}
  \label{fig:illu}
  \vspace*{-3ex}
\end{SCfigure}

To generate the labels for screw displacements, we consider one of the objects, $o_i$, as the base object, and calculate the screw displacements between temporally displaced poses of the second object $o_j$ with respect to it, as illustrated in Fig.~\ref{fig:illu}. Given a sequence of $n$ images $\mathcal{I}_{1:n}$, we calculate a sequence of $n-1$ screw displacements $^1\boldsymbol{\sigma}_{o_j} =\{^1\sigma_2,...^1\sigma_{n} \}$, where each $^1\sigma_k$ corresponds to the relative spatial displacement between the pose of the object $o_j$ in the first image $\mathcal{I}_1$ and the images $\mathcal{I}_{k,~k\in \{2...n\}}$. Note $^1\boldsymbol{\sigma}_{o_j}$ is defined in the frame $\mathcal{F}_{o_j^1}$ attached to the pose of the object $o_j$ in the first image $\mathcal{I}_1$. We can transform $^1\boldsymbol{\sigma}_{o_j}$ to a frame attached to the base object $\mathcal{F}_{o_i}$ by defining the 3D line motion matrix $\tilde{D}$ (Eqn.~\ref{eq:2}) between the frames $\mathcal{F}_{o^1_j}$ and $\mathcal{F}_{o_i}$ \cite{bartoli20013d}, and transforming the common screw axis $\mathsf{^1S}$ to the target frame $\mathcal{F}_{o_i}$. The configurations $^1q_k$ remain the same during frame transformations.


\section{Experiments}
\label{sec:experiments}
\vspace*{-3pt}
We evaluated ScrewNet's performance in estimating the articulation models for objects by conducting three sets of experiments on two benchmarking datasets: the simulated articulated objects dataset provided by Abbatematteo et al. \cite{abbatematteo2019learning} and the recently proposed PartNet-Mobility dataset \cite{Xiang_2020_SAPIEN, Mo_2019_CVPR, chang2015shapenet}. The first set of experiments evaluated ScrewNet's performance in estimating the articulation models for unseen object instances that belong to the object classes used for training the network. Next, we tested ScrewNet's performance in estimating the model parameters for novel articulated objects that belong to the same articulation model category as seen during training. In the third set of experiments, we trained a single ScrewNet on object instances belonging to different object classes and articulation model categories and evaluated its performance in cross-category articulation model estimation. We compared ScrewNet with a state-of-the-art articulation model estimation method proposed by Abbatematteo et al. \cite{abbatematteo2019learning}. Lastly, to evaluate how effectively ScrewNet transfers from simulation to real-world setting, we trained ScrewNet solely using simulated images and tested it to estimate articulation models for three real-world objects. 


In all the experiments, we assumed that the input depth images are semantically segmented and contain non-zero pixels corresponding only to the two objects between which we wish to estimate the articulation model. Given this input, ScrewNet estimates the articulation model parameters for the pair of objects in an object-centric coordinate frame defined at the center of the bounding box of the object.
Note while the approach proposed by Abbatematteo et al. \cite{abbatematteo2019learning} can be used to estimate the articulation model parameters directly in the camera frame, for a fair comparison to our approach, we modified the baseline to predict the model parameters in the object-centric reference frame as well. 

\def\figurename{Fig.}

\begin{figure*}
\centering
    \includegraphics[trim=25 20 25 25, clip, width=\textwidth, height=0.18\textwidth]{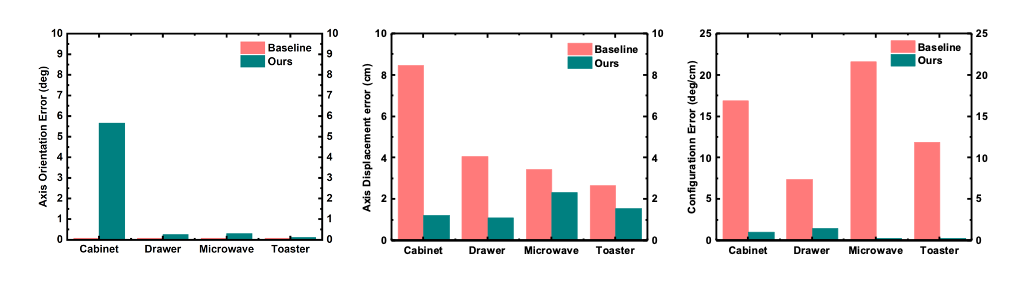}
    \includegraphics[trim=25 20 25 25, clip, width=\textwidth, height=0.18\textwidth]{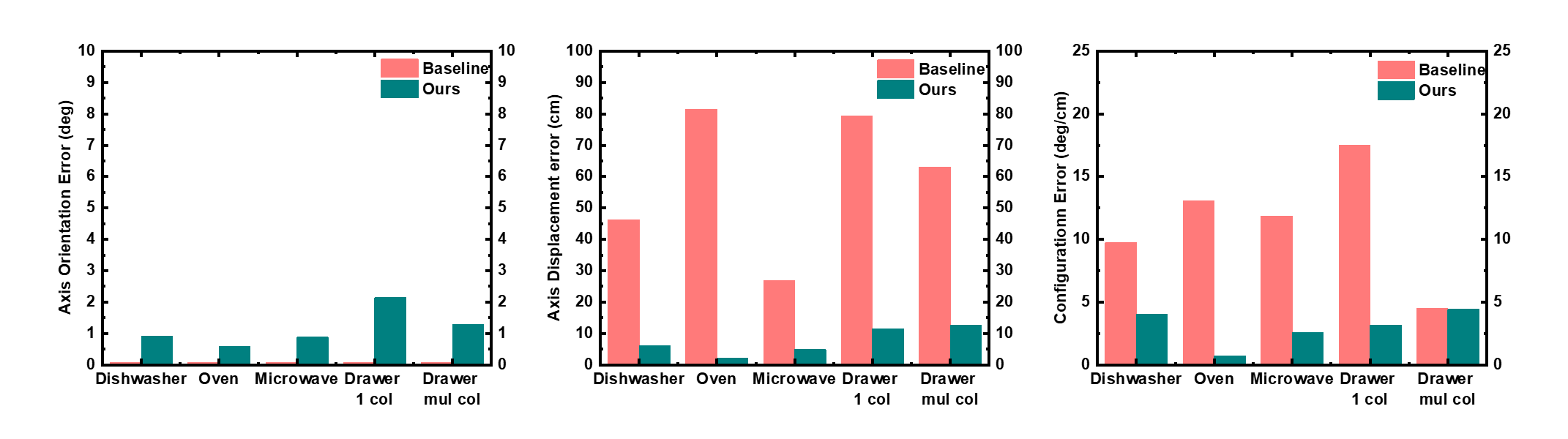}
    \vspace*{-4.5ex}
    \caption{\textbf{[Same object class]} Mean error values for the joint axis orientations, positions, and joint configurations for 1000 test objects for each object class from \textbf{(top)} the simulated articulated objects dataset \cite{abbatematteo2019learning} and \textbf{(bottom)} PartNet-Mobility Dataset \cite{Xiang_2020_SAPIEN, Mo_2019_CVPR, chang2015shapenet}. Configuration errors for all drawers are in cm the remaining configuration errors are in degrees.}
    \label{fig:exp1}
\end{figure*}
\begin{figure*}
\centering
    \includegraphics[trim=25 30 25 25, clip, width=\textwidth, height=0.18\textwidth]{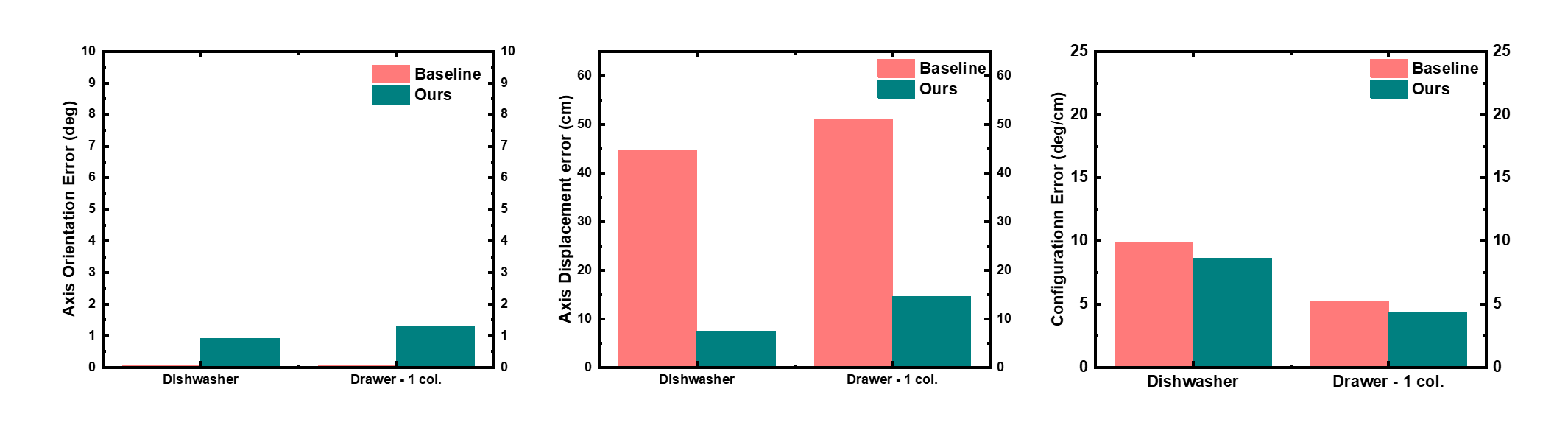}
    \vspace*{-4.5ex}
    \caption{\textbf{[Same articulation model category]} Mean errors for the joint axis orientations, positions, and joint configurations for 1000 test objects for each object class  from the PartNet-Mobility Dataset.}
    \label{fig:exp3}
  \vspace*{-3ex}
\end{figure*}

\subsection{Same object class}
\vspace*{-3pt}
In the first set of experiments, we investigated whether our proposed approach can generalize to unseen object instances belonging to the object classes seen during the training. For this set of experiments, we trained a separate ScrewNet and a baseline network \cite{abbatematteo2019learning} for each of the object classes and tested how ScrewNet fares in comparison to the baseline under similar experimental conditions. We generated 10,000 training examples for each object class in both datasets and performed evaluations on 1,000 withheld object instances. From Fig.~\ref{fig:exp1}, it is evident that ScrewNet outperformed the baseline in estimating the joint axis position and the observed joint configurations by a significant margin for the first dataset. However, for the joint axis orientation estimation, the baseline method reported lower errors than the ScrewNet. Similar trends in the performance of the two methods were observed on the PartNet-Mobility dataset (see Fig.~\ref{fig:exp1}). ScrewNet significantly outperformed the baseline method in estimating the joint axis displacement and observed joint configurations, while the baseline reported lower errors than ScrewNet in estimating the joint axis orientations. However, for both the datasets, the errors reported by ScrewNet in screw axis orientation estimation were reasonably low ($<\ang{5}$), and the model parameters predicted by ScrewNet may be used directly for manipulating the object. These experiments demonstrate that under similar experimental conditions, ScrewNet can estimate the joint axis positions and joint configurations for objects better than the baseline method, while reporting reasonably low but higher errors in joint axis orientations.


\subsection{Same articulation model category}
\begin{figure*}
\centering
    \includegraphics[trim=25 30 25 25, clip, width=\textwidth, height=0.18\textwidth]{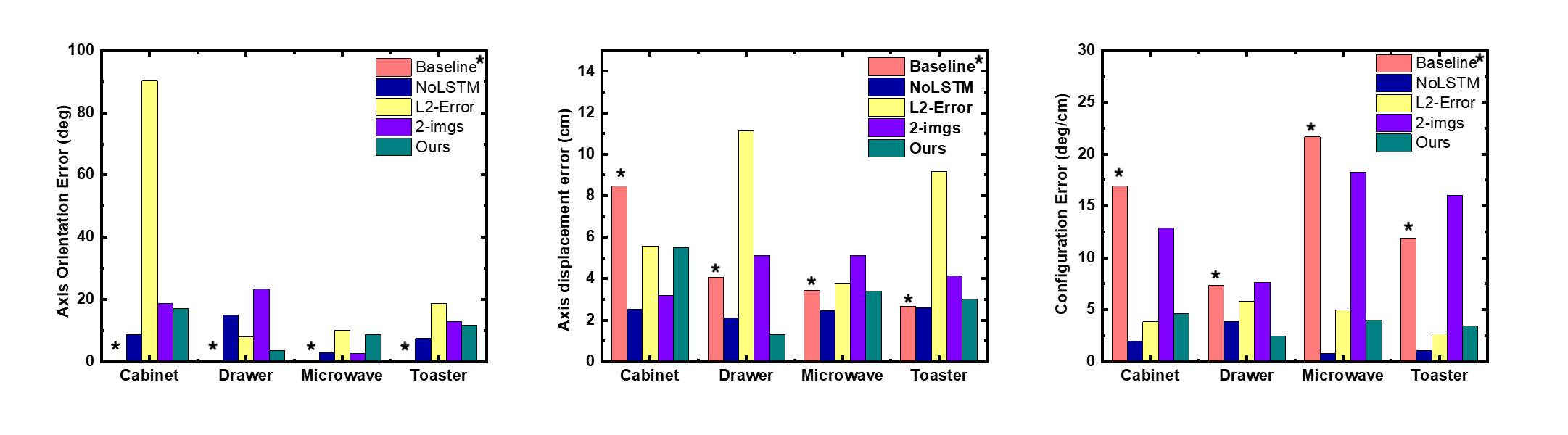}
    \includegraphics[trim=25 30 25 25, clip, width=\textwidth, height=0.18\textwidth]{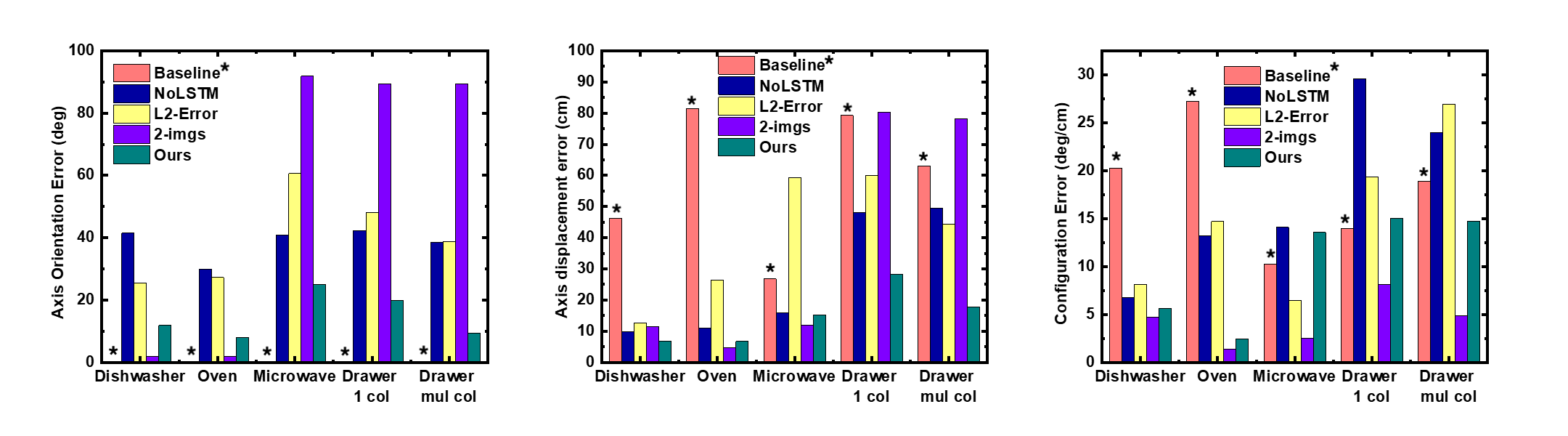}
    \vspace*{-4.5ex}
    \caption{\textbf{[Across articulation model category]} Mean error values for the joint axis orientations, positions, and joint configurations for 1000 test objects for each object class from \textbf{(top)} the simulated articulated objects dataset \cite{abbatematteo2019learning} and \textbf{(bottom)} PartNet-Mobility Dataset \cite{Xiang_2020_SAPIEN, Mo_2019_CVPR, chang2015shapenet}. Symbol $^\star$ denote that the baseline has a significant advantage over other methods as it uses a separate network for each object class}
    \label{fig:exp4}
  \vspace*{-3ex}
\end{figure*}

\vspace*{-3pt}
Next, we investigated if our proposed approach can generalize to unseen object classes belonging to the same articulation model category. We conducted this set of experiments only on the PartNet-Mobility dataset as the simulated articulated objects dataset does not contain enough variety of object classes belonging to the same articulation model category (only 3 for revolute and 1 for prismatic). For the revolute category, we trained ScrewNet and the baseline on the object instances generated from the oven and the microwave object classes and tested it on the objects from the dishwasher class. For the prismatic category, we trained them on the objects from the storage furniture class containing multiples columns of drawers and tested it on the storage furniture objects containing a single column of drawers. We trained a single instance of ScrewNet and the baseline for each articulation model category and used them to predict the articulation model parameters for the test object classes. We used the same training datasets as used in the previous set of experiments. Results are reported in Fig.~\ref{fig:exp3}. It is evident from Fig.~\ref{fig:exp4} that ScrewNet was able to generalize to novel object classes belonging to the same articulation model category, while the baseline failed to do so. Both methods reported low errors in the joint axis orientation and the observed configurations. However, for the joint axis position, the baseline method reported mean errors of an order of magnitude higher than the ScrewNet for both the articulation model categories. 
\subsection{Across articulation model category}
Next, we studied whether ScrewNet can estimate articulation model parameters for unseen objects across the articulation model category. For these experiments, we trained a single ScrewNet on a mixed dataset consisting of object instances belonging to all object model classes. To test whether sharing training data across articulation categories can help in reducing the number of examples required for training, we used only half of the dataset available for each object class ($5000$ examples each) while preparing the mixed dataset. We compared its performance with a baseline network that is trained specifically on the particular object class. We also conducted ablation studies to test the effectiveness of the various components of the proposed method.

Fig.~\ref{fig:exp4} summarizes the results for the first dataset. Even though we used a single network to estimate the articulation model for objects belonging to different articulation model categories, ScrewNet performed at par or better than the baseline method for all the object model categories. ScrewNet outperformed the baseline while estimating the observed joint configurations for all object classes, even though the baseline was trained separately for each object class. For joint axis position estimation, ScrewNet reported significantly lower errors than the baseline for the cabinet and the drawer classes, and comparable errors for the microwave and the toaster classes. In estimating the joint axis orientations, both methods reported comparable errors for the cabinet, drawer, and the toaster classes. However, for the cabinet object class, ScrewNet reported a higher error than the baseline method, which may stem from the fact that the cabinet object class includes both left-opening and right-opening configurations that have a difference $\ang{180}$ in their axis orientations. On the PartNet-Mobility dataset (see Fig.~\ref{fig:exp4}), the performances of the methods was similar, with ScrewNet outperforming the baseline method with a significant margin in estimating the joint axis positions and the observed joint configurations while reporting higher errors than the baseline in estimating the joint axis orientations. The results show ScrewNet leverages the unified representation and performs cross-category articulation model estimation with better on average performance than the current state-of-the-art method while using only half the training examples.


In comparison to its ablated versions, ScrewNet outperformed the L2-error and the two-images versions by a significant margin for both datasets and performed comparably to the NoLSTM version. For the first dataset, the NoLSTM version reported lower errors than ScrewNet in estimating the joint axis orientations, their positions, and the observed joint configurations for the microwave, cabinet, and toaster classes. However, the NoLSTM version failed to generalize across articulation model categories and reported higher errors than the ScrewNet for the drawer class, and sometimes even predicted NaNs. On the second dataset, ScrewNet reported much lower errors than the NoLSTM ablated version for all object model categories. These results demonstrate that for reliably estimating articulation model parameters across categories, both the sequential information available in the input and a loss function that grounds predictions in their physical meaning are crucial.

\begin{figure}[b]
\centering
    \includegraphics[width=0.48\textwidth, height=0.11\textwidth]{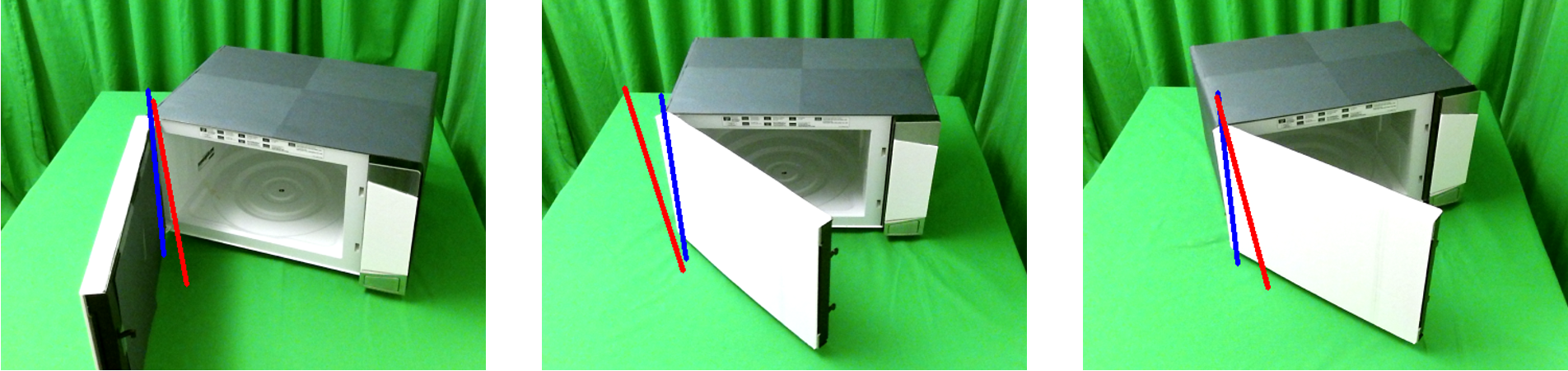}
    \vspace*{-4.5ex}
    \caption{\textbf{[Real-world images]} Images with overlayed ground-truth joint axis (blue) and predicted axis (red) for different poses of the microwave}
    \label{fig:exp7}
  \vspace*{-3ex}
\end{figure}

\subsection{Real world images}
Finally, we evaluated how effectively ScrewNet transfers from simulation to a real-world setting. ScrewNet was trained solely on the combined simulated articulated object dataset. Afterwards, we used the model to infer the joint axis of a microwave, a drawer, and a toaster oven. Figure~\ref{fig:exp7} qualitatively demonstrates ScrewNet's performance for three different poses of the microwave. Despite only ever having seen simulated data, ScrewNet achieved a mean error of $\sim\ang{20}$ in axis orientation and $\sim 1.5\text{cm}$ in axis position on real-world sensor input. These results demonstrate that ScrewNet achieves reasonable estimates of the articulation model parameters for real-world objects when it is trained solely using simulated data. In order to obtain better performances a retraining on real world data would be required.

\section{Conclusion} 
\label{sec:conclusion}
\vspace*{-5pt}
Articulated objects are common in human environments and service robots will be interacting with them frequently while assisting humans. For manipulating such objects, a robot will need to learn the articulation properties of such objects through raw sensory data such as RGB-D images. Current methods for estimating the articulation model of objects from visual observations either require textured objects or need to know the articulation model category a priori for estimating the articulation model parameters from the depth images. We propose ScrewNet that uses screw theory to unify the representation of different articulation models and performs category-independent articulation model estimation from depth images. We evaluate the performance of ScrewNet on two benchmarking datasets and compare it with a state-of-the-art method. Results show that ScrewNet can estimate articulation models and their parameters for objects across object classes and articulation model categories successfully with better on average performance than the baselines while using half the training data and without requiring to know the model category.\footnote{For further details and results, refer: \href{https://arxiv.org/abs/2008.10518}{https://arxiv.org/abs/2008.10518}}

While ScrewNet successfully performs cross-category articulation model estimation, at present, it can only predict 1-DOF articulation models. For multi-DoF objects, an image segmentation step is required to mask out all non-relevant object parts. This procedure can be repeated iteratively on all object part pairs to estimate relative models between object parts. Future work will directly estimate multi-DoF objects by learning a segmentation network along with the ScrewNet. Further work work will predict the articulation model parameters directly in the robot's camera frame rather than in an object-centric frame. Predictions in the camera frame can help the robot to learn articulation models in an active learning fashion.

\section{Acknowledgement}
This work has taken place in the Personal Autonomous Robotics Lab (PeARL) at The University of Texas at Austin. PeARL research is supported in part by the NSF (IIS-1724157, IIS-1638107, IIS-1749204, IIS-1925082), ONR (N00014-18-2243), AFOSR (FA9550-20-1-0077), and ARO (78372-CS).  This research was also sponsored by the Army Research Office under Cooperative Agreement Number W911NF-19-2-0333. The views and conclusions contained in this document are those of the authors and should not be interpreted as representing the official policies, either expressed or implied, of the Army Research Office or the U.S. Government. The U.S. Government is authorized to reproduce and distribute reprints for Government purposes notwithstanding any copyright notation herein. The authors would also like to thank Wonjoon Goo, Yuchen Cui, and Akanksha Saran for their insightful discussions and Tushti Shah for her help with graphics.

\printbibliography

@inproceedings{abbatematteo2019learning,
  title={Learning to Generalize Kinematic Models to Novel Objects},
  author={Abbatematteo, Ben and Tellex, Stefanie and Konidaris, George},
  booktitle={Proceedings of the Third Conference on Robot Learning},
  year={2019}
}

@book{siciliano2016springer,
  title={Springer handbook of robotics},
  author={Siciliano, Bruno and Khatib, Oussama},
  year={2016},
  publisher={Springer}
}

@book{mason2001mechanics,
  title={Mechanics of robotic manipulation},
  author={Mason, Matthew T},
  year={2001},
  publisher={MIT press}
}

@article{jia2019,
author = {Jia, Yan-bin},
pages = {1--11},
title = {\href{http://web.cs.iastate.edu/~cs577/handouts/plucker-coordinates.pdf}{Pl{\"{u}}cker Coordinates for Lines in the Space [Lecture Notes]}},
month= {August},
year = {2019},
publisher={Iowa State University}
}

@inproceedings{li2020category,
  title={Category-Level Articulated Object Pose Estimation},
  author={Li, Xiaolong and Wang, He and Yi, Li and Guibas, Leonidas J and Abbott, A Lynn and Song, Shuran},
  booktitle={Proceedings of the IEEE/CVF Conference on Computer Vision and Pattern Recognition},
  pages={3706--3715},
  year={2020}
}

@article{martin2019coupled,
  title={Coupled recursive estimation for online interactive perception of articulated objects},
  author={Mart{\'{i}}n-Mart{\'{i}}n, Roberto and Brock, Oliver},
  journal={The International Journal of Robotics Research},
  pages={0278364919848850},
  year={2019},
  publisher={SAGE Publications Sage UK: London, England}
}

@inproceedings{sturm2009learning,
  title={Learning Kinematic Models for Articulated Objects.},
  author={Sturm, J{\"u}rgen and Pradeep, Vijay and Stachniss, Cyrill and Plagemann, Christian and Konolige, Kurt and Burgard, Wolfram},
  booktitle={IJCAI},
  pages={1851--1856},
  year={2009}
}

@article{sturm2011probabilistic,
  title={A probabilistic framework for learning kinematic models of articulated objects},
  author={Sturm, J{\"u}rgen and Stachniss, Cyrill and Burgard, Wolfram},
  journal={Journal of Artificial Intelligence Research},
  volume={41},
  pages={477--526},
  year={2011}
}

@article{pillai2015learning,
  title={Learning articulated motions from visual demonstration},
  author={Pillai, Sudeep and Walter, Matthew R and Teller, Seth},
  journal={arXiv preprint arXiv:1502.01659},
  year={2015}
}

@inproceedings{niekum2015online,
  title={Online bayesian changepoint detection for articulated motion models},
  author={Niekum, Scott and Osentoski, Sarah and Atkeson, Christopher G and Barto, Andrew G},
  booktitle={2015 IEEE International Conference on Robotics and Automation (ICRA)},
  pages={1468--1475},
  year={2015},
  organization={IEEE}
}

@article{jain2019learning,
  title={Learning hybrid object kinematics for efficient hierarchical planning under uncertainty},
  author={Jain, Ajinkya and Niekum, Scott},
  journal={IEEE International Conference on Intelligent Robots and Systems (IROS)},
  year={2020}
}

@inproceedings{katz2013interactive,
  title={Interactive segmentation, tracking, and kinematic modeling of unknown 3d articulated objects},
  author={Katz, Dov and Kazemi, Moslem and Bagnell, J Andrew and Stentz, Anthony},
  booktitle={2013 IEEE International Conference on Robotics and Automation},
  pages={5003--5010},
  year={2013},
  organization={IEEE}
}

@inproceedings{katz2008manipulating,
  title={Manipulating articulated objects with interactive perception},
  author={Katz, Dov and Brock, Oliver},
  booktitle={2008 IEEE International Conference on Robotics and Automation},
  pages={272--277},
  year={2008},
  organization={IEEE}
}

@inproceedings{martin2014online,
  title={Online interactive perception of articulated objects with multi-level recursive estimation based on task-specific priors},
  author={Mart{\'{i}}n-Mart{\'{i}}n, Roberto and Brock, Oliver},
  booktitle={2014 IEEE/RSJ International Conference on Intelligent Robots and Systems},
  pages={2494--2501},
  year={2014},
  organization={IEEE}
}

@inproceedings{bartoli20013d,
  title={The 3D line motion matrix and alignment of line reconstructions},
  author={Bartoli, Adrien and Sturm, Peter},
  booktitle={Proceedings of the 2001 IEEE Computer Society Conference on Computer Vision and Pattern Recognition. CVPR 2001},
  volume={1},
  pages={I--I},
  year={2001},
  organization={IEEE}
}

@inproceedings{perez2017c,
  title={C-learn: Learning geometric constraints from demonstrations for multi-step manipulation in shared autonomy},
  author={P{\'e}rez-D'Arpino, Claudia and Shah, Julie A},
  booktitle={Robotics and Automation (ICRA), 2017 IEEE International Conference on},
  pages={4058--4065},
  year={2017},
  organization={IEEE}
}

@article{kroemer2019review,
  title={A Review of Robot Learning for Manipulation: Challenges, Representations, and Algorithms},
  author={Kroemer, Oliver and Niekum, Scott and Konidaris, George},
  journal={arXiv preprint arXiv:1907.03146},
  year={2019}
}

@inproceedings{jain2018efficient,
  title={Efficient hierarchical robot motion planning under uncertainty and hybrid dynamics},
  author={Jain, Ajinkya and Niekum, Scott},
  booktitle={Conference on Robot Learning},
  pages={757--766},
  year={2018}
}

@incollection{daniele2020multiview,
  title={A multiview approach to learning articulated motion models},
  author={Daniele, Andrea F and Howard, Thomas M and Walter, Matthew R},
  booktitle={Robotics Research},
  pages={371--386},
  year={2020},
  publisher={Springer}
}

@article{liu2019learning,
  title={Learning Articulated Constraints From a One-Shot Demonstration for Robot Manipulation Planning},
  author={Liu, Yizhou and Zha, Fusheng and Sun, Lining and Li, Jingxuan and Li, Mantian and Wang, Xin},
  journal={IEEE Access},
  volume={7},
  pages={172584--172596},
  year={2019},
  publisher={IEEE}
}

@inproceedings{desingh2019factored,
  title={Factored pose estimation of articulated objects using efficient nonparametric belief propagation},
  author={Desingh, Karthik and Lu, Shiyang and Opipari, Anthony and Jenkins, Odest Chadwicke},
  booktitle={2019 International Conference on Robotics and Automation (ICRA)},
  pages={7221--7227},
  year={2019},
  organization={IEEE}
}

@inproceedings{Yi2018,
author = {Yi, Li and Huang, Haibin and Liu, Difan and Kalogerakis, Evangelos and Su, Hao and Guibas, Leonidas},
booktitle = {SIGGRAPH Asia 2018 Tech. Pap. SIGGRAPH Asia 2018},
doi = {10.1145/3272127.3275027},
eprint = {1809.07417},
title = {{Deep part induction from articulated object pairs}},
url = {https://dl.acm.org/doi/10.1145/3272127.3275027},
volume = {37},
year = {2018}
}

@inproceedings{michel2015pose,
  title={Pose Estimation of Kinematic Chain Instances via Object Coordinate Regression.},
  author={Michel, Frank and Krull, Alexander and Brachmann, Eric and Yang, Michael Ying and Gumhold, Stefan and Rother, Carsten},
  booktitle={BMVC},
  pages={181--1},
  year={2015}
}

@InProceedings{Xiang_2020_SAPIEN,
author = {Xiang, Fanbo and Qin, Yuzhe and Mo, Kaichun and Xia, Yikuan and Zhu, Hao and Liu, Fangchen and Liu, Minghua and Jiang, Hanxiao and Yuan, Yifu and Wang, He and Yi, Li and Chang, Angel X. and Guibas, Leonidas J. and Su, Hao},
title = {{SAPIEN}: A SimulAted Part-based Interactive ENvironment},
booktitle = {The IEEE Conference on Computer Vision and Pattern Recognition (CVPR)},
month = {June},
year = {2020}}

@InProceedings{Mo_2019_CVPR,
author = {Mo, Kaichun and Zhu, Shilin and Chang, Angel X. and Yi, Li and Tripathi, Subarna and Guibas, Leonidas J. and Su, Hao},
title = {{PartNet}: A Large-Scale Benchmark for Fine-Grained and Hierarchical Part-Level {3D} Object Understanding},
booktitle = {The IEEE Conference on Computer Vision and Pattern Recognition (CVPR)},
month = {June},
year = {2019}
}

@article{chang2015shapenet,
title={Shapenet: An information-rich 3d model repository},
author={Chang, Angel X and Funkhouser, Thomas and Guibas, Leonidas and Hanrahan, Pat and Huang, Qixing and Li, Zimo and Savarese, Silvio and Savva, Manolis and Song, Shuran and Su, Hao and others},
journal={arXiv preprint arXiv:1512.03012},
year={2015}
}

@inproceedings{hausman2015active,
  title={Active articulation model estimation through interactive perception},
  author={Hausman, Karol and Niekum, Scott and Osentoski, Sarah and Sukhatme, Gaurav S},
  booktitle={Robotics and Automation (ICRA), 2015 IEEE International Conference on},
  pages={3305--3312},
  year={2015},
  organization={IEEE}
}

@inproceedings{jain2009pulling,
  title={Pulling open novel doors and drawers with equilibrium point control},
  author={Jain, Advait and Kemp, Charles C},
  booktitle={Humanoid Robots, 2009. Humanoids 2009. 9th IEEE-RAS International Conference on},
  pages={498--505},
  year={2009},
  organization={IEEE}
}

@article{bohg2017interactive,
  title={Interactive perception: Leveraging action in perception and perception in action},
  author={Bohg, Jeannette and Hausman, Karol and Sankaran, Bharath and Brock, Oliver and Kragic, Danica and Schaal, Stefan and Sukhatme, Gaurav S},
  journal={IEEE Transactions on Robotics},
  volume={33},
  number={6},
  pages={1273--1291},
  year={2017},
  publisher={IEEE}
}

@inproceedings{baum2017opening,
  title={Opening a lockbox through physical exploration},
  author={Baum, Manuel and Bernstein, Matthew and Mart{\'{I}}n-Mart{\'{I}}n, Roberto and H{\"o}fer, Sebastian and Kulick, Johannes and Toussaint, Marc and Kacelnik, Alex and Brock, Oliver},
  booktitle={2017 IEEE-RAS 17th International Conference on Humanoid Robotics (Humanoids)},
  pages={461--467},
  year={2017},
  organization={IEEE}
}

@inproceedings{martin2016integrated,
  title={An integrated approach to visual perception of articulated objects},
  author={Mart{\'{i}}n-Mart{\'{i}}n, Roberto and H{\"o}fer, Sebastian and Brock, Oliver},
  booktitle={2016 IEEE International Conference on Robotics and Automation (ICRA)},
  pages={5091--5097},
  year={2016},
  organization={IEEE}
}

@article{bochkovskiy2020yolov4,
  title={YOLOv4: Optimal Speed and Accuracy of Object Detection},
  author={Bochkovskiy, Alexey and Wang, Chien-Yao and Liao, Hong-Yuan Mark},
  journal={arXiv preprint arXiv:2004.10934},
  year={2020}
}

@inproceedings{he2016deep,
  title={Deep residual learning for image recognition},
  author={He, Kaiming and Zhang, Xiangyu and Ren, Shaoqing and Sun, Jian},
  booktitle={Proceedings of the IEEE conference on computer vision and pattern recognition},
  pages={770--778},
  year={2016}
}

@inproceedings{todorov2012mujoco,
  title={Mujoco: A physics engine for model-based control},
  author={Todorov, Emanuel and Erez, Tom and Tassa, Yuval},
  booktitle={2012 IEEE/RSJ International Conference on Intelligent Robots and Systems},
  pages={5026--5033},
  year={2012},
  organization={IEEE}
}

@article{gupta2019relay,
  title={Relay Policy Learning: Solving Long-Horizon Tasks via Imitation and Reinforcement Learning},
  author={Gupta, Abhishek and Kumar, Vikash and Lynch, Corey and Levine, Sergey and Hausman, Karol},
  journal={arXiv preprint arXiv:1910.11956},
  year={2019}
}
\flushcolsend

\clearpage
\section*{Appendix}
\subsection{Experimental details}
\subsubsection{Dataset}
\label{appendix:1}
\begin{figure}[t]
    \centering
    \includegraphics[width=0.6\linewidth]{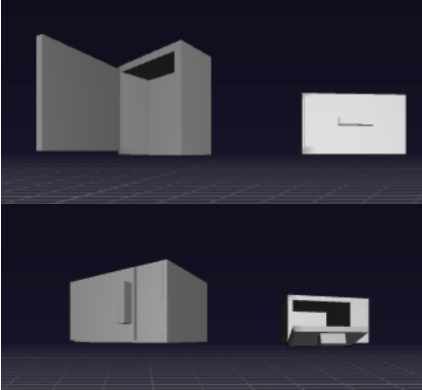}
    \caption{Object classes used from the simulated articulated object dataset \cite{abbatematteo2019learning}. Object classes: cabinet, drawer, microwave, and toaster (left to right)}
    \label{fig:simArt-dataset}
\end{figure}

\begin{figure}[t]
    \centering
    \includegraphics[width=0.7\linewidth]{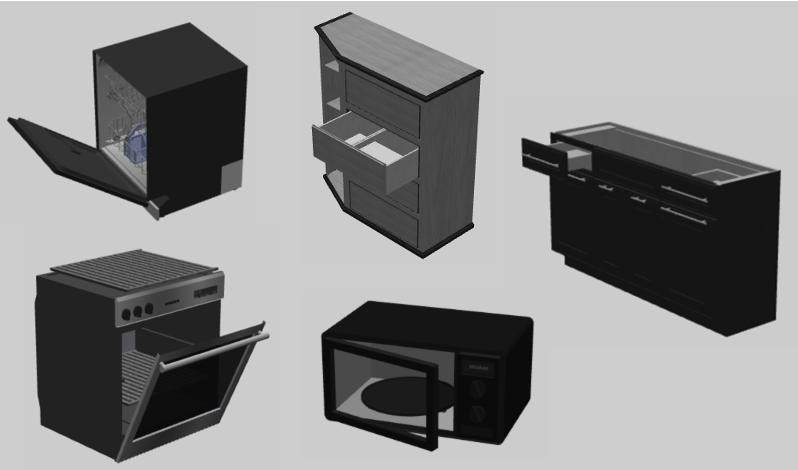}
    \caption{Object classes used from the PartNet-Mobility dataset \cite{Xiang_2020_SAPIEN, Mo_2019_CVPR, chang2015shapenet}. Object classes: dishwasher, oven, microwave, drawer- 1 column, and drawer- multiple columns (left to right)}
    \label{fig:partNet-dataset}
\end{figure}

Objects used in the experiments from each of the dataset are shown in the Figures~\ref{fig:simArt-dataset} and \ref{fig:partNet-dataset}. We sampled a new object geometry and a joint location for each training example in the simulated articulated object dataset, as proposed by \cite{abbatematteo2019learning}. For the PartNet-Mobility dataset, we considered $11$ microwave ($8$ train, $3$ test), $36$ dishwasher ($27$ train, $9$ test), $9$ oven ($6$ train, $3$ test), $26$ single column drawer ($20$ train, $6$ test), and $14$ multi-column drawer ($10$ train, $4$ test) object models. For both datasets, we sampled object positions and orientations uniformly in the view frustum of the camera up to a maximum depth dependent upon the object size.

\subsubsection{Experiment 1: Same object class}
Numerical error values for the first set of experiments for the simulated articulated objects dataset are presented in the Table~\ref{table:table1}. It is evident from the Table~~\ref{table:table1} that the baseline succeeded in achieving nearly zero prediction error ($\ang{0.08}$) in joint axis orientation estimation for all object classes. ScrewNet also performed well and reported low prediction errors ($<\ang{0.5}$) for the drawer, microwave, and toaster object classes. For the cabinet object class, while ScrewNet reported a higher mean error ($\sim\ang{5}$), it is relatively small compared to the difference in axis orientations, $\ang{180}$, between the two possible configurations of the cabinet (left-opening or right-opening). For the other two model parameters, namely the joint axis position and the observed configurations, ScrewNet significantly outperformed the baseline method.

\begingroup
\setlength{\tabcolsep}{6pt} 
\renewcommand{\arraystretch}{1.25} 
\begin{table*}[h]
\centering
\resizebox{\textwidth}{!}{%
\begin{tabular}{cccc}
\hline
                     & \textbf{Axis Orientation (deg)} & \textbf{Axis displacement (cm)} & \textbf{Configuration} \\ \hline
Cabinet - Baseline   & $\mathbf{0.082 \pm 0.000}$  & $8.472 \pm 6.277$ & $16.921 \pm 8.212$ deg \\
Cabinet - Ours       & $5.667 \pm 13.888$ & $\mathbf{1.236 \pm 0.66}$  & $\mathbf{1.083 \pm 0.707}$ deg \\ \hline
Drawer - Baseline    & $\bm{0.082 \pm 0.000}$  & $4.067 \pm 1.483$ & $7.389 \pm 2.439$ cm   \\
Drawer - Ours        & $0.252 \pm 0.000$  & $\bm{1.114 \pm 0.035}$ & $\bm{1.517 \pm 0.016}$  cm  \\ \hline
Microwave - Baseline & $\bm{0.083 \pm 0.000}$  & $3.441 \pm 1.218$ & $21.670 \pm 7.097$ deg \\
Microwave - Ours     & $0.304 \pm 0.000$  & $\bm{2.322 \pm 1.323}$ & $\bm{0.329 \pm 0.005}$ deg  \\ \hline
Toaster - Baseline   & $\bm{0.082 \pm 0.000}$  & $2.669 \pm 1.338$ & $11.902 \pm 4.242$ deg \\
Toaster - Ours       & $0.114 \pm 0.000$   & $\bm{1.566 \pm 0.018}$ & $\bm{0.314 \pm 0.018}$ deg  \\ \hline
\end{tabular}%
}
\caption{Mean error values for joint axis orientation, joint axis position, and configurations for 1000 test object instances for each object class from the simulated articulated objects dataset \cite{abbatematteo2019learning}. Lowest error values for a particular test object set are reported in bold.}
\label{table:table1}
\end{table*}
\endgroup

Numerical error values for the first set of experiments for the PartNet-Mobility dataset are reported in the Table~\ref{table:partNet1}. Similar trends followed in the performance of the two approaches. The baseline achieved very high accuracy in predicting the joint axis orientation, whereas ScrewNet reported reasonably low but slightly higher errors ($<\ang{2.5}$). For the joint axis position and the observed configurations, ScrewNet outperformed the baseline method on this dataset as well.

\begingroup
\setlength{\tabcolsep}{6pt} 
\renewcommand{\arraystretch}{1.25} 
\begin{table*}
\centering
\resizebox{\textwidth}{!}{%
\begin{tabular}{cccc}
\hline
                                & \textbf{Axis Orientation (deg)} & \textbf{Axis displacement (cm)} & \textbf{Configuration} \\ \hline
Dishwasher - Baseline & $\bm{0.082 \pm 0.000}$ & $46.267 \pm 20.247$ & $9.735 \pm 4.6$ deg \\
Dishwasher - Ours & $0.918 \pm 0.000$ & $\bm{6.136 \pm 5.455}$ & $\bm{4.037 \pm 1.613}$ deg \\ \hline
Oven - Baseline & $\bm{0.082 \pm 0.000}$ & $81.444 \pm 27.083$ & $13.087 \pm 4.649$  deg \\
Oven - Ours & $0.583 \pm 0.223$ & $\bm{2.111 \pm 1.910}$ & $\bm{0.720 \pm 0.140}$   deg \\ \hline
Microwave - Baseline & $\bm{0.082 \pm 0.000}$ & $26.781 \pm 10.273$ & $11.856 \pm 2.456$ \\
Microwave - Ours & $0.879 \pm 0.063$ & $\bm{4.893 \pm 4.252}$ & $\bm{2.549 \pm 0.939}$ deg \\ \hline
Drawer- 1 column - Baseline & $\bm{0.082 \pm 0.000}$ & $79.228 \pm 13.944$ & $17.524 \pm 3.700$ cm \\
Drawer- 1 column - Ours & $2.140 \pm 0.000$ & $\bm{11.567 \pm 9.748}$ & $\bm{3.181 \pm 0.793}$  cm \\ \hline
Drawer- Multi. cols. - Baseline & $\bm{0.082 \pm 0.000}$ & $63.064 \pm 18.913$ & $4.483 \pm 6.403$ cm \\
Drawer- Multi. cols. - Ours & $1.287 \pm 0.000$ & $\bm{12.557 \pm 8.317}$ & $\bm{4.419 \pm 2.891}$ cm \\ \hline
\end{tabular}%
}
\caption{Mean error values for joint axis orientation, joint axis position, and configurations for 1000 test cases for each object class from the PartNet-Mobility Dataset}
\label{table:partNet1}
\end{table*}
\endgroup

\subsubsection{Experiment 2: Same articulation model category}
Numerical results for the second set of experiments are reported in the Table~\ref{table:partNet-same-art}. It is evident from the Table~\ref{table:partNet-same-art} that the ScrewNet was able to generalize to novel object classes belonging to the same articulation model category, while the baseline method failed to do so. While both approaches reported comparable errors in estimating the joint axis orientations and the observed configurations, the baseline reported errors of an order of magnitude higher than ScrewNet in the joint axis position estimation.

\begingroup
\setlength{\tabcolsep}{6pt} 
\renewcommand{\arraystretch}{1.25} 
\begin{table*}[h]
\centering
\resizebox{\textwidth}{!}{%
\begin{tabular}{cccc}
\hline
                                    & \textbf{Axis Orientation (deg)} & \textbf{Axis displacement (cm)} & \textbf{Configuration} \\ \hline
Oven - Baseline & $\bm{0.082 \pm 0.000}$ & $44.699 \pm 12.259$ & $9.915 \pm 3.934$ deg \\
Oven - Ours & $0.918 \pm 0.000$ & $\bm{7.486 \pm 1.273}$ & $\bm{8.650 \pm 0.207}$ deg \\ \hline
Drawer- 1 column - Baseline & $\bm{0.082 \pm 0.000}$ & $50.990 \pm 25.984$ & $5.283 \pm 8.862$ cm \\
Drawer- 1 column - Ours & $1.287 \pm 0.000$ & $\bm{14.548 \pm 5.823}$ & $\bm{4.399 \pm 0.654}$ cm \\ \hline
\end{tabular}%
}
\caption{Mean error values for joint axis orientation, joint axis position, and configurations for 1000 test objects belonging to each object classes from the PartNet-Mobility Dataset}
\label{table:partNet-same-art}
\end{table*}
\endgroup

\subsubsection{Ablation studies}
\label{appendix:2}
We consider three ablated versions of ScrewNet. First, to test the effectiveness of the proposed loss function, we consider an ablated version of ScrewNet which is trained using a raw L2-loss between the labels and the network predictions (named as L2-Error version while reporting results).
As the second ablation study, we test whether using an LSTM layer in the network helps with the performance or not (named as NoLSTM version while reporting results). We replace the LSTM layer of the ScrewNet with a fully connected layer such that the two networks, ScrewNet and its ablated version, have a comparable number of parameters. Lastly, to check if a sequence of images is helpful in the model estimation or not, we consider an ablated version of ScrewNet that estimates the articulation model using just a pair of images (named as 2\_imgs version while reporting results). Note that ScrewNet and all its ablated versions use a single network each. Numerical results for the simulated articulated objects dataset are presented in the Table~\ref{table:table3}, and for the PartNet-Mobility dataset are shown in the Table~\ref{table:partNet3}.

\begingroup
\setlength{\tabcolsep}{6pt} 
\renewcommand{\arraystretch}{1.25} 
\begin{table*}[h]
\centering
\resizebox{\textwidth}{!}{%
\begin{tabular}{cccc}
\hline
                     & \textbf{Axis Orientation (deg)} & \textbf{Axis displacement (cm)} & \textbf{Configuration} \\ \hline
Cabinet - Baseline$^\star$ & $\bm{0.082 \pm 0.000}$ & $8.472 \pm 6.277$ & $16.921 \pm 8.212$ deg \\
Cabinet - NoLSTM & $8.688 \pm 20.504$ & $\bm{2.521 \pm 4.341}$ & $\bm{1.984 \pm 5.172}$ deg \\
Cabinet - L2-Error & $90.186 \pm 12.244$ & $5.580 \pm 5.138$ & $3.847 \pm 5.377$ deg \\
\multicolumn{1}{l}{Cabinet - 2\_imgs} & $18.716 \pm 40.197$ & $3.188 \pm 5.795$ & $12.898 \pm 8.846$ deg \\
Cabinet - Ours & $16.988 \pm 14.971$ & $5.479 \pm 4.363$ & $4.65 \pm 5.904$ deg \\ \hline
Drawer - Baseline$^\star$ & $\bm{0.082 \pm 0.000}$ & $4.067 \pm 1.483$ & $7.389 \pm 2.439$ cm \\
Drawer - NoLSTM & $14.957 \pm 25.526$ & $2.116 \pm 2.287$ & $3.878 \pm 2.883$ cm \\
Drawer - L2-Error & $7.931 \pm 13.745$ & $11.141 \pm 3.159$ & $5.847 \pm 1.468$ cm \\
\multicolumn{1}{l}{Drawer - 2\_imgs} & $23.310 \pm 27.888$ & $5.118 \pm 2.829$ & $7.664 \pm 4.883$ cm \\
Drawer - Ours & $3.473 \pm 8.839$ & $\bm{1.302 \pm 0.999}$ & $\bm{2.448 \pm 1.092}$ cm \\ \hline
Microwave - Baseline$^\star$ & $\bm{0.082 \pm 0.000}$ & $3.441 \pm 1.218$ & $21.67 \pm 7.097$ deg \\
Microwave - NoLSTM & $2.725 \pm 8.813$ & $\bm{2.439 \pm 1.708}$ & $\bm{0.803 \pm 2.519}$ deg \\
Microwave - L2-Error & $10.125 \pm 10.953$ & $3.76 \pm 3.021$ & $4.957 \pm 4.489$ deg \\
\multicolumn{1}{l}{Microwave - 2\_imgs} & $2.547 \pm 3.480$ & $5.115 \pm 5.076$ & $18.269 \pm 12.658$ deg \\
Microwave - Ours & $8.770 \pm 13.363$ & $3.398 \pm 2.675$ & $4.033 \pm 5.998$ deg \\ \hline
Toaster - Baseline$^\star$ & $\bm{0.082 \pm 0.000}$ & $2.669 \pm 1.338$ & $11.902 \pm 4.242$ deg \\
Toaster - NoLSTM & $7.410 \pm 17.645$ & $\bm{2.597 \pm 1.86}$ & $\bm{1.030 \pm 2.230}$ deg \\
Toaster - L2-Error & $18.750 \pm 17.243$ & $9.173 \pm 4.229$ & $2.661 \pm 2.823$ deg \\
\multicolumn{1}{l}{Toaster - 2\_imgs} & $12.833 \pm 22.596$ & $4.123 \pm 3.196$ & $16.016 \pm 10.703$ deg \\
Toaster - Ours & $11.583 \pm 14.798$ & $3.003 \pm 1.75$ & $3.471 \pm 2.876$ deg \\ \hline
\end{tabular}%
}
\caption{Mean error values for joint axis orientation, joint axis position, and configurations for 1000 test objects belonging to each object classes from the simulated articulated objects dataset. Symbol $^\star$ denote that the baseline has a significant advantage over other methods as it uses a separate network for each object class, while all ScrewNet and its ablations use a single network}
\label{table:table3}
\end{table*}
\endgroup

\begingroup
\setlength{\tabcolsep}{6pt} 
\renewcommand{\arraystretch}{1.25} 
\begin{table*}[h]
\centering
\resizebox{\textwidth}{!}{%
\begin{tabular}{cccc}
\hline
 & \textbf{Axis Orientation (deg)} & \textbf{Axis displacement (cm)} & \textbf{Configuration} \\ \hline
Dishwasher - Baseline$^\star$ & $\bm{0.082 \pm 0.000}$ & $46.267 \pm 20.247$ & $9.735 \pm 4.600$ deg \\
Dishwasher - NoLSTM & $41.485 \pm 41.184$ & $9.815 \pm 6.782$ & $\bm{5.415 \pm 4.097}$ deg \\
Dishwasher - L2 Error & $25.405 \pm 15.119$ & $12.653 \pm 8.119$ & $7.828 \pm 1.913$ deg \\
Dishwasher - 2\_imgs & $1.935 \pm 0.021$ & $11.544 \pm 4.729$ & $5.706 \pm 4.152$ deg \\
Dishwasher - Ours & $11.850 \pm 15.267$ & $\bm{6.789 \pm 5.630}$ & $6.081 \pm 3.043$ deg \\ \hline
Oven - Baseline$^\star$ & $\bm{0.082 \pm 0.000}$ & $81.429 \pm 27.244$ & $13.026 \pm 4.670$ deg \\
Oven - NoLSTM & $29.968 \pm 39.034$ & $11.014 \pm 13.235$ & $10.574 \pm 6.332$ deg \\
Oven - L2 Error & $27.197 \pm 13.103$ & $26.452 \pm 14.704$ & $11.823 \pm 1.067$ deg \\
Oven - 2\_imgs & $1.939 \pm 0.018$ & $\bm{4.791 \pm 1.370}$ & $10.498 \pm 7.481$ deg \\
Oven - Ours & $7.881 \pm 7.763$ & $6.786 \pm 2.443$ & $\bm{5.010 \pm 1.233}$ deg \\ \hline
Microwave - Baseline$^\star$ & $\bm{0.082 \pm 0.000}$ & $26.781 \pm 10.273$ & $11.856 \pm 2.456$ deg \\
Microwave - NoLSTM & $40.911 \pm 32.830$ & $15.993 \pm 14.080$ & $3.865 \pm 2.350$ deg \\
Microwave - L2 Error & $60.566 \pm 7.705$ & $59.286 \pm 6.485$ & $7.463 \pm 1.612$ deg \\
Microwave - 2\_imgs & $91.826 \pm 0.012$ & $\bm{11.994 \pm 2.549}$ & $5.212 \pm 3.606$ deg \\
Microwave - Ours & $24.959 \pm 24.847$ & $15.271 \pm 13.561$ & $\bm{3.507 \pm 1.987}$ deg \\ \hline
Drawer- 1 col. - Baseline$^\star$ & $\bm{0.082 \pm 0.000}$ & $79.228 \pm 13.944$ & $17.524 \pm 3.700$ cm \\
Drawer- 1 col. - NoLSTM & $42.318 \pm 35.604$ & $47.991 \pm 29.586$ & $10.923 \pm 6.449$ cm \\
Drawer- 1 col. - L2 Error & $48.136 \pm 9.533$ & $60.046 \pm 19.375$ & $14.202 \pm 2.153$ cm \\
Drawer - 1 col. - 2\_imgs & $89.372 \pm 0.047$ & $80.356 \pm 8.087$ & $25.753 \pm 18.374$ cm \\
Drawer- 1 col. - Ours & $19.876 \pm 21.684$ & $\bm{28.329 \pm 15.005}$ & $\bm{5.729 \pm 4.259}$ cm \\ \hline
Drawer- Multi. cols. - Baseline$^\star$ & $\bm{0.082 \pm 0.000}$ & $63.064 \pm 18.913$ & $4.483 \pm 6.403$ cm \\
Drawer- Multi. cols.- NoLSTM & $38.393 \pm 33.113$ & $49.419 \pm 23.998$ & $6.181 \pm 5.228$ cm \\
Drawer- Multi. cols.- L2 Error & $38.866 \pm 5.243$ & $44.422 \pm 26.927$ & $6.422 \pm 0.766$ cm \\
Drawer- Multi. cols. - 2\_imgs & $89.361 \pm 0.053$ & $78.131 \pm 4.888$ & $12.229 \pm 3.961$ cm \\
Drawer- Multi. cols. - Ours & $9.292 \pm 15.295$ & $\bm{17.813 \pm 14.719}$ & $\bm{0.915 \pm 1.772}$ cm \\ \hline
\end{tabular}%
}
\caption{[Experiment: Across articulation model category] Mean error values for joint axis orientation, joint axis position, and configurations for 1000 test objects belonging to each object classes from the PartNet-Mobility Dataset. Symbol $^\star$ denote that the baseline has a significant advantage over other methods as it uses a separate network for each object class, while all ScrewNet and its ablations use a single network}
\label{table:partNet3}
\end{table*}
\endgroup

\end{document}